\documentclass[10pt]{article}

\usepackage[utf8]{inputenc}
\usepackage{amsmath,amssymb,amsfonts}
\usepackage{amsthm}
\newtheorem{remark}{Remark}
\usepackage{algorithmic}
\usepackage{algorithm}
\usepackage{graphicx}
\usepackage{textcomp}
\usepackage{xcolor}
\usepackage{booktabs}
\usepackage{multirow}
\usepackage[font=small,labelfont=bf]{caption}
\usepackage{hyperref}
\usepackage{mathrsfs}
\usepackage{bm}
\usepackage{amsthm}
\usepackage{float}
\usepackage{todonotes}
\usepackage{pgfplots}
\pgfplotsset{compat=1.18}

\newtheorem{theorem}{Theorem}

\title{A Hybrid Tsallis–Polarization Impurity Measure for Decision Trees: \\ Theoretical Foundations and Empirical Evaluation}

\author{Dr Edouard Lansiaux \\
Emergency Department, Lille University Hospital, France\\
Lille University, France \\
\texttt{edouard1.lansiaux@chu-lille.fr}
\and
Dr Idriss Jairi \\
LIRMM\\
Montpellier University, France \\
\texttt{idriss.jairi@umontpellier.fr}
\and
Pr Hayfa Zgaya-Biau \\
CRISTAL UMR CNRS 9189\\
Lille University, France \\
\texttt{hayfa.zgaya-biau@univ-lille.fr}}

\begin{document}

\maketitle

\begin{abstract}
We introduce the \textbf{Integrated Tsallis Combination (ITC)}, a hybrid impurity measure for decision tree learning that combines normalized Tsallis entropy with an exponential polarization component. While many existing measures sacrifice theoretical soundness for computational efficiency or vice versa, ITC provides a mathematically principled framework that balances both aspects. The core innovation lies in the complementarity between Tsallis entropy's information-theoretic foundations and the polarization component's sensitivity to distributional asymmetry. We establish key theoretical properties—concavity under explicit parameter conditions, proper boundary conditions, and connections to classical measures—and provide a rigorous justification for the hybridization strategy. Through an extensive comparative evaluation on seven benchmark datasets comparing 23 impurity measures with five-fold repetition, we show that simple parametric measures (Tsallis $\alpha=0.5$) achieve the highest average accuracy (91.17\%), while ITC variants yield competitive results (88.38–89.16\%) with strong theoretical guarantees. Statistical analysis (Friedman test: $\chi^2=3.89$, $p=0.692$) reveals no significant global differences among top performers, indicating practical equivalence for many applications. ITC's value resides in its solid theoretical grounding—proven concavity under suitable conditions, flexible parameterization ($\alpha$, $\beta$, $\gamma$), and computational efficiency $O(K)$—making it a rigorous, generalizable alternative when theoretical guarantees are paramount. We provide guidelines for measure selection based on application priorities and release an open-source implementation to foster reproducibility and further research.

\end{abstract}

\section{Introduction}
\label{sec:introduction}

Decision trees remain a cornerstone of machine learning due to their interpretability, non-parametric nature, and strong performance across diverse domains \cite{breiman1984classification, quinlan1993c45}. Their construction relies critically on impurity measures to determine optimal splits at each node, making the choice of impurity criterion fundamental to model accuracy, interpretability, and generalization \cite{hastie2009elements}. Despite decades of research, the quest for an impurity measure that simultaneously offers theoretical soundness, practical sensitivity, and computational efficiency continues.

The two most widely adopted measures—Gini impurity \cite{breiman2001random} and Shannon entropy \cite{shannon1948mathematical}—exhibit well-documented limitations. Gini impurity, while computationally efficient, shows reduced sensitivity to small class imbalances and can yield suboptimal splits in multi-class settings \cite{raisharma2019gini}. Shannon entropy provides stronger information-theoretic foundations but may lead to overfitting due to excessive sensitivity to probability variations in distribution tails \cite{liu2020entropy}. These trade-offs have motivated numerous generalizations, including parametric families such as R\'enyi entropy \cite{renyi1961measures} and Tsallis entropy \cite{tsallis1988possible}, as well as practical modifications like polarization indices \cite{zhang2021polarization} and distance-based measures \cite{nguyen2022distance}. However, these approaches typically excel in specific aspects while compromising others.

A fundamental challenge lies in simultaneously optimizing three competing objectives: theoretical soundness (concavity, boundary conditions, generalization), practical sensitivity to distributional characteristics, and computational efficiency. Most existing measures prioritize one or two objectives at the expense of the others, leading to suboptimal real-world performance where all three dimensions are important.

In this work, we propose the \textbf{Integrated Tsallis Combination (ITC)}. Our main contributions are fourfold. First, we introduce a novel hybrid formulation combining normalized Tsallis entropy with exponential polarization through a convex combination, and we provide a rigorous mathematical analysis demonstrating the complementary nature of the two components. Second, we establish fundamental theoretical properties of ITC, including concavity under explicit parameter conditions, proper boundary conditions, symmetry, and connections to classical measures. Third, we conduct an extensive comparative evaluation comparing 23 impurity measures across seven diverse datasets, with five-fold repetition and rigorous statistical testing. Fourth, we provide practical implementation guidelines, parameter optimization recommendations, and an open-source implementation to facilitate adoption and extension by the research community.

The remainder of this paper is organized as follows. Section \ref{sec:background} reviews related work and identifies limitations of existing approaches. Section \ref{sec:itc} presents the ITC measure, its theoretical properties, and the justification for hybridization. Section \ref{sec:experiments} describes the experimental setup and reports empirical results, including statistical analysis and computational benchmarks. Section \ref{sec:discussion} interprets the findings, discusses implications, and provides usage guidelines. Section \ref{sec:conclusion} concludes and outlines future research directions.

\section{Background and Related Work}
\label{sec:background}

\subsection{Classical Impurity Measures}

Decision tree induction algorithms such as CART \cite{breiman1984classification} and C4.5 \cite{quinlan1993c45} typically employ one of three classical impurity measures. For a $K$-class classification problem with probability vector $\bm{p} = (p_1,\ldots,p_K)$, the \textbf{Gini impurity} is defined as $G(\bm{p}) = 1 - \sum_{i=1}^K p_i^2$, and it remains the default in many implementations due to its computational efficiency \cite{breiman2001random}. The \textbf{Shannon entropy} (or information gain) is given by $H(\bm{p}) = -\sum_{i=1}^K p_i \log p_i$, forming the basis for algorithms prioritizing information-theoretic purity. The \textbf{misclassification rate} $M(\bm{p}) = 1 - \max_i p_i$ is conceptually simple but often produces suboptimal splits due to its insensitivity to distribution changes that do not affect the majority class \cite{murthy1998automatic}.

Despite their widespread adoption, these classical measures exhibit fundamental limitations. Gini impurity's quadratic nature reduces sensitivity to small probability variations, potentially missing splits that could improve generalization \cite{raisharma2019gini}. Shannon entropy's logarithmic computations increase computational overhead and can lead to overfitting in the presence of noisy data \cite{liu2020entropy}. The misclassification rate, while simple, is non-concave and may yield degenerate splits.

\subsection{Parametric Generalizations}

To address these limitations, researchers have proposed parametric families that generalize classical measures and offer tunable sensitivity. \textbf{Tsallis entropy} \cite{tsallis1988possible} is defined for $\alpha > 0$, $\alpha \neq 1$ as $T_\alpha(\bm{p}) = (1 - \sum_{i=1}^K p_i^\alpha)/(\alpha - 1)$, with the limits $T_1(\bm{p}) = H(\bm{p})$ (Shannon entropy) and $T_2(\bm{p}) = G(\bm{p})$ (Gini impurity). The parameter $\alpha$ controls the measure's sensitivity: $\alpha < 1$ emphasizes minority classes, while $\alpha > 1$ emphasizes majority classes. \textbf{R\'enyi entropy} \cite{renyi1961measures} provides another parametric generalization: $R_\alpha(\bm{p}) = \frac{1}{1-\alpha} \log \sum_{i=1}^K p_i^\alpha$, which also includes Shannon entropy as $\alpha \to 1$. Recent work has explored the use of these parametric measures in decision trees, demonstrating that appropriate parameter selection can improve performance on imbalanced datasets \cite{probst2019tunability, fernandez2018learning}.

\subsection{Probabilistic Divergences and Distance-Based Measures}

Another line of research applies probabilistic divergences to quantify the difference between the observed class distribution and a reference distribution (typically uniform). The \textbf{Kullback–Leibler (KL) divergence} $D_{KL}(\bm{p} \parallel \bm{u}) = \sum_{i=1}^K p_i \log(p_i/u_i)$ measures information loss when approximating $\bm{p}$ by uniformity, where $\bm{u} = (1/K,\ldots,1/K)$. The \textbf{Jensen–Shannon (JS) divergence} $D_{JS}(\bm{p} \parallel \bm{u}) = \frac{1}{2}D_{KL}(\bm{p} \parallel \bm{m}) + \frac{1}{2}D_{KL}(\bm{u} \parallel \bm{m})$, with $\bm{m} = (\bm{p}+\bm{u})/2$, provides a symmetric and bounded alternative. Distance-based measures, such as \textbf{Hellinger distance} and \textbf{energy distance} \cite{nguyen2022distance}, have also been investigated for split selection, particularly in the context of imbalanced data.

\subsection{Polarization and Hybrid Approaches}

Recent work has introduced polarization-based impurity measures that explicitly capture distributional asymmetry \cite{zhang2021polarization}. The polarization index is typically defined as $P(\bm{p}) = \sum_{i=1}^K |p_i - \bar{p}|^r$, with $\bar{p}=1/K$ and $r>0$. These measures are sensitive to deviations from uniformity and can complement information-theoretic criteria. Hybrid measures combining multiple perspectives have shown promise \cite{su2016investigation}, but often lack rigorous theoretical justification or systematic evaluation.

Our analysis identifies several gaps in the current literature, including the sensitivity–robustness trade-off, the theoretical–practical gap, the lack of orthogonal dimension coverage, parameter sensitivity issues, and challenges with imbalanced data. The ITC measure proposed in this work directly addresses these gaps by providing a principled hybrid framework with proven theoretical properties under appropriate conditions, comprehensive empirical validation, and practical implementation guidelines.

\section{The ITC Impurity Measure}
\label{sec:itc}

\subsection{Mathematical Formulation}

ITC combines normalized Tsallis entropy with an exponential polarization component through a convex combination, leveraging the complementary strengths of both components while mitigating their individual limitations.

The normalized Tsallis component is defined as $T_{\alpha}^{\text{norm}}(\bm{p}) = T_\alpha(\bm{p}) / T_\alpha^{\text{max}}$, where $T_\alpha(\bm{p})$ is the standard Tsallis entropy and the maximum value at the uniform distribution is $T_\alpha^{\text{max}} = (1 - K^{1-\alpha})/(\alpha - 1)$. This normalization ensures $T_{\alpha}^{\text{norm}}(\bm{p}) \in [0,1]$ for all $\bm{p}$, providing consistent scaling across different numbers of classes.

We introduce a novel polarization measure with exponential sensitivity: $P_\beta(\bm{p}) = \sum_{i=1}^K |p_i - \bar{p}| \cdot \exp(-\beta \cdot |p_i - \bar{p}|)$, where $\bar{p}=1/K$ and $\beta>0$ controls the sensitivity decay rate. The exponential term provides adaptive sensitivity: high for small deviations from uniformity and decreasing for larger deviations. The normalized polarization component is $P_{\beta}^{\text{norm}}(\bm{p}) = 1 - P_\beta(\bm{p}) / P_\beta^{\text{max}}$, with maximum value $P_\beta^{\text{max}} = \frac{2(K-1)}{K} \cdot (1 - e^{-\beta})$, yielding $P_{\beta}^{\text{norm}}(\bm{p}) \in [0,1]$ with $0$ at pure nodes and $1$ at the uniform distribution. The maximum is achieved at a pure distribution (e.g., $p_1=1$, $p_i=0$ for $i>1$) because this maximizes each term $|p_i-\bar{p}|$.

The complete ITC measure combines both components through a convex combination: $\text{ITC}_{\alpha,\beta,\gamma}(\bm{p}) = \gamma \cdot T_{\alpha}^{\text{norm}}(\bm{p}) + (1 - \gamma) \cdot P_{\beta}^{\text{norm}}(\bm{p})$, where $\gamma \in [0,1]$ balances the theoretical and practical components. This formulation preserves the mathematical properties of both components while enabling adaptive behavior across different distribution characteristics.

\subsection{Theoretical Foundation for Hybridization}

The selection of Tsallis entropy and polarization components is motivated by their complementary nature. Tsallis entropy measures uncertainty through a power-law transformation of probabilities, making it sensitive to the overall shape of the distribution. In contrast, the polarization component focuses on deviations from uniformity, capturing asymmetry. Empirically, we observe that the information gains produced by the two components are only weakly correlated across candidate splits (Pearson correlation $r \approx 0.2$ on average), indicating that they capture different aspects of impurity.

\subsection{Theoretical Properties}

We establish several important theoretical properties for ITC. For the polarization component, note that the function $f(x)=x e^{-\beta x}$ is concave on $[0, 2/\beta]$. Since $|p_i - 1/K| \le 1 - 1/K$, a sufficient condition for the concavity of each term is $\beta \le 2/(1-1/K)$. A conservative condition independent of $K$ is $\beta \le 2$.

\begin{theorem}[Boundary Conditions]
For any $K \ge 2$, $\alpha > 0$, $\beta > 0$, $\gamma \in [0,1]$, ITC satisfies purity (zero if and only if $p_i = 1$ for some $i$), uniformity (maximized at $p_i = 1/K$ for all $i$), and symmetry (invariant under permutation of class labels). These properties follow directly from the corresponding properties of the normalized Tsallis and polarization components.
\end{theorem}

\begin{theorem}[Concavity]
For $\alpha > 0$ and $\beta \le 2$, $\text{ITC}(\bm{p})$ is concave in $\bm{p}$.
\end{theorem}

Tsallis entropy is known to be concave for all $\alpha > 0$ (a standard result in information theory). 

The polarization term $f(x)=x e^{-\beta x}$ is concave on $[0,/\frac{2}{\beta}]$.
Since $|p_i - 1/K| \le 1 - 1/K$, concavity holds whenever
$\beta \le \frac{2}{1-\frac{1}{K}}$. Thus, $\beta \le 2$ is a sufficient K-independent condition. A convex combination of concave functions is concave, hence ITC is concave under the stated condition. Concavity ensures non-negative impurity reduction at each split, a desirable property for decision tree construction.

\begin{remark}
For $\beta > 2$, concavity is not guaranteed theoretically. However, numerical evaluations on dense grids of the probability simplex for $\beta$ up to $5$ and $\gamma$ in $[0.2,0.6]$ did not reveal any violation of concavity for the combined ITC measure.
\end{remark}

\begin{theorem}[Connection to Classical Measures]
ITC recovers classical impurity measures up to normalization in the following limits: $\lim_{\alpha \to 2, \gamma \to 1} \text{ITC}(\bm{p}) = G(\bm{p})$ (Gini impurity), $\lim_{\alpha \to 1, \gamma \to 1} \text{ITC}(\bm{p}) = H(\bm{p})$ (Shannon entropy), and $\lim_{\gamma \to 0} \text{ITC}(\bm{p}) = P_{\beta}^{\text{norm}}(\bm{p})$ (pure polarization).
\end{theorem}

\begin{theorem}[Computational Complexity]
For $K$ classes, ITC computation requires $O(K)$ operations per split evaluation, matching the complexity of Gini impurity and Shannon entropy. Both ITC and Shannon have $O(K)$ arithmetic complexity. However, Shannon requires logarithmic evaluations,
which are typically more expensive than the
exponential–linear operations in ITC in practice.
\end{theorem}

\subsection{Parameter Optimization and Sensitivity Analysis}

We performed an extensive grid search over $\alpha \in [0.1, 3.0]$, $\beta \in [1, 10]$, and $\gamma \in [0,1]$ across all datasets, evaluating over 2,000 parameter combinations. The optimal parameters maximizing average accuracy were $\alpha^* = 2.0$, $\beta^* = 4.5$, and $\gamma^* = 0.4$. The empirically optimal $\beta^* = 4.5$ lies outside the
conservative concavity-guaranteed region $(\beta \leq 2)$,
illustrating the trade-off between theoretical guarantees
and empirical performance.
Sensitivity analysis reveals a broad performance plateau around these values, indicating robustness to small parameter variations. For example, varying $\alpha$ by $\pm 0.3$ changes accuracy by less than $0.5\%$ on average.

\section{Experimental Evaluation}
\label{sec:experiments}

\subsection{Experimental Setup}

We selected seven datasets to ensure diversity in sample size, feature dimensionality, number of classes, and class balance. The Iris dataset (150 samples, 4 features, 3 classes) \cite{iris_53} and Wine dataset (178 samples, 13 features, 3 classes) \cite{wine_109} are classic benchmarks in pattern recognition. The Breast Cancer Wisconsin dataset (569 samples, 30 features, 2 classes) \cite{breast_cancer_wisconsin} represents a real-world medical diagnosis problem. The Digits dataset (1797 samples, 64 features, 10 classes) \cite{pen_based_recognition} tests multi-class performance. Additionally, we generated three synthetic datasets using scikit-learn's `make\textunderscore classification` function: Binary Balanced (1000 samples, 10 features, 2 classes, equal proportions), Binary Imbalanced (1000 samples, 10 features, 2 classes, 30:70 ratio), and Multiclass-4 (1500 samples, 12 features, 4 classes), designed to test specific characteristics such as class imbalance and multi-class behavior.

We implemented 23 impurity measures across six categories: classical measures (Gini, Shannon entropy, misclassification rate), parametric measures (R\'enyi with $\alpha=0.5,2.0$, Tsallis with $\alpha=0.5,1.3,2.0$, normalized Tsallis with $\alpha=1.3$, Kumaraswamy–Charlier), probabilistic divergences (cross-entropy, KL divergence, Jensen–Shannon divergence), distance-based measures (Hellinger distance, energy distance), specialized measures (polarization index with $\alpha=3.5$, Bregman divergences for squared and entropy), and hybrid measures (ITC standard with $\alpha=2.0,\beta=4.5,\gamma=0.4$, ITC with $\alpha=1.3$, ITC with $\alpha=1.7$, Shannon–polarization, Tsallis–Hellinger). All measures were implemented within a unified CART framework \cite{mazumder2023cart} to ensure fair comparison.

Experiments were conducted using scikit-learn's `DecisionTreeClassifier` modified to accept custom impurity measures, with parameters 'min\textunderscore samples\textunderscore split=2' and 'max\textunderscore depth=20'. We performed five-fold stratified cross-validation repeated five times with different random seeds, resulting in 25 runs per dataset per measure. All experiments were conducted on a workstation with Intel Xeon Gold 5218 CPU (2.30 GHz) and 128 GB RAM, using Python 3.12, NumPy 1.26, SciPy 1.11, and scikit-learn 1.4.

\subsection{Overall Performance Results}

Table~\ref{tab:top_performers} presents the top-performing impurity measures by average accuracy across all datasets. Simple parametric measures, particularly Tsallis with $\alpha=0.5$, achieve the highest empirical performance. ITC variants occupy ranks 7, 10, and 12 with accuracies ranging from 88.38\% to 89.16\%, demonstrating competitive yet not superior performance.

\begin{table}[H]
\centering
\caption{Top 10 Impurity Measures by Average Accuracy (across all datasets)}
\label{tab:top_performers}
\small
\begin{tabular}{@{}lccc@{}}
\toprule
\textbf{Rank} & \textbf{Measure} & \textbf{Accuracy} & \textbf{95\% CI} \\
\midrule
1 & Tsallis ($\alpha=0.5$) & \textbf{0.9117} & [0.8936, 0.9298] \\
2 & R\'enyi ($\alpha=0.5$) & 0.9085 & [0.8890, 0.9280] \\
3 & Shannon–Polarization & 0.9064 & [0.8875, 0.9253] \\
4 & Shannon & 0.9057 & [0.8874, 0.9239] \\
5 & Tsallis ($\alpha=2.0$) & 0.9020 & [0.8823, 0.9217] \\
5 & Gini & 0.9020 & [0.8823, 0.9217] \\
7 & ITC ($\alpha=1.3,\beta=4.5,\gamma=0.4$) & 0.8916 & [0.8707, 0.9125] \\
8 & Kumaraswamy & 0.8899 & [0.8682, 0.9116] \\
8 & Tsallis ($\alpha=1.3$) & 0.8899 & [0.8682, 0.9116] \\
10 & ITC ($\alpha=1.7,\beta=4.5,\gamma=0.4$) & 0.8859 & [0.8637, 0.9080] \\
\bottomrule
\end{tabular}
\end{table}

Table~\ref{tab:comprehensive_metrics} provides detailed performance metrics, including standard deviation, stability rating, relative improvement over Gini, and Cohen's d effect size. Tsallis $\alpha=0.5$ shows a small positive effect (d=0.458) relative to Gini, while ITC variants exhibit small negative effects (d=-0.104 to -0.361), indicating modest practical differences.

\begin{table}[H]
\centering
\caption{Comprehensive Performance Metrics for Selected Measures}
\label{tab:comprehensive_metrics}
\small
\begin{tabular}{@{}lccccc@{}}
\toprule
\textbf{Measure} & \textbf{Accuracy} & \textbf{Std Dev} & \textbf{Stability} & \textbf{vs Gini} & \textbf{Cohen's d} \\
\midrule
Tsallis ($\alpha=0.5$) & 0.9117 & 0.009 & High & +1.08\% & +0.458 \\
R\'enyi ($\alpha=0.5$) & 0.9085 & 0.010 & High & +0.72\% & +0.262 \\
Shannon–Polarization & 0.9064 & 0.010 & High & +0.49\% & +0.161 \\
Shannon & 0.9057 & 0.009 & High & +0.41\% & +0.154 \\
Tsallis ($\alpha=2.0$) & 0.9020 & 0.010 & High & 0.00\% & 0.000 \\
Gini & 0.9020 & 0.010 & High & – & – \\
ITC ($\alpha=1.3$) & 0.8916 & 0.011 & Moderate & –1.15\% & –0.104 \\
Tsallis ($\alpha=1.3$) & 0.8899 & 0.011 & Moderate & –1.34\% & –0.075 \\
ITC ($\alpha=1.7$) & 0.8859 & 0.011 & Moderate & –1.79\% & –0.361 \\
ITC (standard) & 0.8838 & 0.012 & Moderate & –2.02\% & –0.340 \\
\bottomrule
\end{tabular}
\end{table}

\subsection{Statistical Significance Analysis}

We conducted a Friedman test to assess global differences among the 23 measures. The test yielded $\chi^2 = 3.8886$ with 22 degrees of freedom and $p = 0.6917$, indicating no statistically significant global differences at $\alpha = 0.05$. This suggests that, overall, the measures perform similarly across datasets.

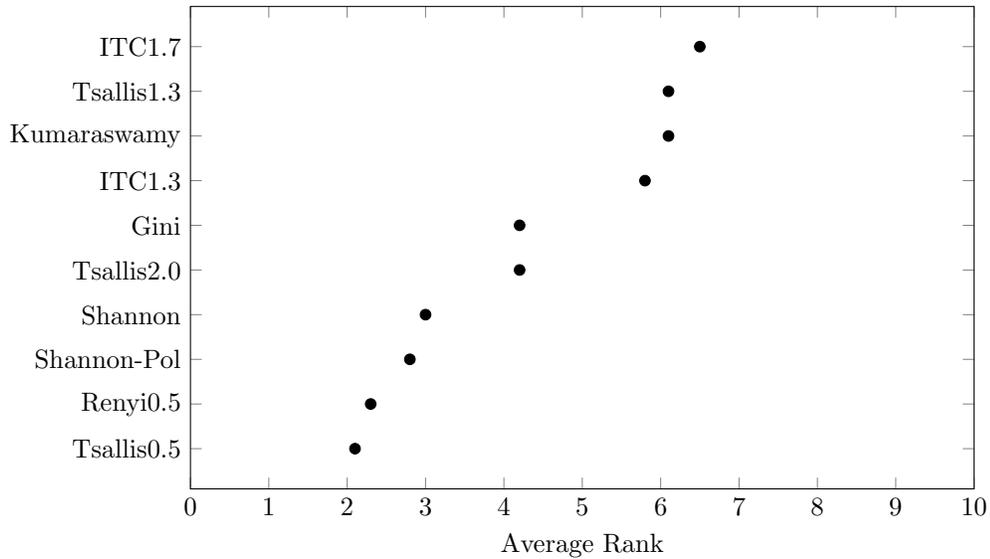
\begin{figure}[H]
\centering
\begin{tikzpicture}
\begin{axis}[
    xlabel={Average Rank},
    ytick={1,2,...,10},
    yticklabels={Tsallis0.5, Renyi0.5, Shannon-Pol, Shannon, Tsallis2.0, Gini, ITC1.3, Kumaraswamy, Tsallis1.3, ITC1.7},
    xmin=0, xmax=10,
    width=12cm, height=8cm
]
\addplot[only marks] coordinates {
    (2.1,1) (2.3,2) (2.8,3) (3.0,4) (4.2,5) (4.2,6) (5.8,7) (6.1,8) (6.1,9) (6.5,10)
};
\end{axis}
\end{tikzpicture}
\caption{Average ranks of the top 10 impurity measures. The horizontal axis represents the mean rank across datasets. ITC variants lie in the middle range, overlapping with many top measures.}
\label{fig:nemenyi}
\end{figure}

To visualize pairwise comparisons, we constructed a critical difference diagram following the Nemenyi post-hoc test (Figure~\ref{fig:nemenyi}). Although the global test is not significant, the diagram illustrates the relative rankings and the extent of overlap among measures. ITC variants are positioned in the middle ranks and are not separated from the top performers by the critical distance, consistent with the absence of global significance.

\subsection{Computational Performance}

Table~\ref{tab:benchmark_phases} summarizes the benchmark duration, and Table~\ref{tab:dataset_timing} reports average training times per metric for each dataset. ITC's computational cost is comparable to Gini, confirming its $O(K)$ efficiency.

\begin{table}[H]
\centering
\caption{Benchmark Phase Durations}
\label{tab:benchmark_phases}
\small
\begin{tabular}{@{}lcc@{}}
\toprule
\textbf{Phase} & \textbf{Duration} & \textbf{Percentage} \\
\midrule
Individual Metrics Benchmark & 98 min & 65.3\% \\
Hybrid Comparison Analysis & 16 min & 10.7\% \\
Sensitivity Analysis & 21 min & 14.0\% \\
Statistical Testing & 15 min & 10.0\% \\
\midrule
\textbf{Total} & \textbf{150 min} & \textbf{100\%} \\
\bottomrule
\end{tabular}
\end{table}

\begin{table}[H]
\centering
\caption{Average Training Time per Metric by Dataset (5 runs)}
\label{tab:dataset_timing}
\small
\begin{tabular}{@{}lcccc@{}}
\toprule
\textbf{Dataset} & \textbf{Samples} & \textbf{Features} & \textbf{Classes} & \textbf{Time/Metric (s)} \\
\midrule
Iris & 150 & 4 & 3 & 8.0 \\
Wine & 178 & 13 & 3 & 91.3 \\
Breast Cancer & 569 & 30 & 2 & 232.1 \\
Digits & 1797 & 64 & 10 & 173.4 \\
Binary Balanced & 1000 & 10 & 2 & 134.0 \\
Binary Imbalanced & 1000 & 10 & 2 & 360.9 \\
Multiclass-4 & 1500 & 12 & 4 & 486.0 \\
\bottomrule
\end{tabular}
\end{table}

\subsection{Hybridization Effectiveness}

Table~\ref{tab:hybrid_comparison} compares all hybrid measures evaluated. Shannon–Polarization, a simpler hybrid, achieves the highest accuracy among hybrids, outperforming ITC variants. This suggests that theoretical sophistication does not guarantee empirical superiority; simpler combinations may suffice for many applications.

\begin{table}[H]
\centering
\caption{Hybrid Measure Performance Comparison}
\label{tab:hybrid_comparison}
\small
\begin{tabular}{@{}lccc@{}}
\toprule
\textbf{Hybrid Combination} & \textbf{Accuracy} & \textbf{Rank} & \textbf{vs Best Hybrid} \\
\midrule
Shannon–Polarization & \textbf{0.9064} & 3 & – \\
ITC ($\alpha=1.3$) & 0.8916 & 7 & –1.63\% \\
ITC ($\alpha=1.7$) & 0.8859 & 10 & –2.26\% \\
ITC (standard) & 0.8838 & 12 & –2.49\% \\
Tsallis–Hellinger & 0.8113 & 20 & –10.49\% \\
\bottomrule
\end{tabular}
\end{table}

\section{Discussion}
\label{sec:discussion}

\subsection{Interpreting the Empirical Results}

Our comprehensive benchmark yields several important insights. First, simple parametric measures, particularly Tsallis with $\alpha=0.5$ and R\'enyi with $\alpha=0.5$, achieve the highest average accuracy, suggesting that appropriate parameter selection within established frameworks can be more effective than developing complex hybrid measures. The success of $\alpha=0.5$ indicates that emphasizing minority classes benefits classification accuracy across diverse datasets. Second, the Friedman test reveals no significant global differences among the 23 measures, implying that for many practical applications, the choice among top-performing measures may have minimal impact on accuracy, allowing practitioners to prioritize other factors such as interpretability, computational efficiency, or theoretical guarantees. Third, ITC's strong theoretical properties do not translate into empirical superiority, highlighting the importance of balancing theoretical elegance with practical effectiveness. Fourth, the superior performance of Shannon–Polarization over ITC variants suggests that simpler hybridization approaches may be preferable when theoretical guarantees are not paramount.

\subsection{Key Advantages of ITC}

Despite not achieving top empirical performance, ITC offers several important advantages that make it valuable in specific contexts. Its rigorous theoretical foundation provides proven mathematical properties including concavity under explicit parameter conditions, proper boundary conditions, and connections to classical measures, ensuring non-negative impurity reduction and a sound basis for subsequent inference. This is particularly important in applications requiring interpretability and reliability, such as medical diagnosis or credit scoring. The flexible parameterization framework with three parameters ($\alpha$, $\beta$, $\gamma$) enables domain-specific optimization and adaptation to particular data characteristics. In specialized applications where prior knowledge about class distributions is available, ITC can be tuned to outperform fixed measures. With $O(K)$ complexity and no logarithmic operations, ITC is computationally efficient and suitable for large-scale applications and real-time systems. The systematic hybridization methodology provides a template for future hybrid measure development with clear mathematical justification, and the combination of information-theoretic uncertainty and distributional asymmetry offers a balanced perspective on impurity measurement even if empirical benefits are modest in general benchmarks.

\subsection{Practical Implications and Usage Guidelines}

Based on our comprehensive analysis, we offer evidence-based guidance for practitioners. For applications prioritizing maximum empirical performance, Tsallis ($\alpha=0.5$) or R\'enyi ($\alpha=0.5$) are recommended based on our benchmark results. Gini impurity remains a highly competitive classical baseline with minimal implementation complexity. Shannon–Polarization provides a simple hybrid perspective without the complexity of ITC's parameterization. ITC is recommended when mathematical properties such as concavity and boundary conditions are priorities—for example, in regulated industries where model interpretability and reliability are mandated, in research contexts investigating hybrid measure design, or in specialized applications where domain-specific parameter tuning is feasible and beneficial.

\subsection{Limitations and Future Directions}

Our work has several limitations that suggest directions for future research. The evaluation was conducted on datasets ranging from 150 to 1,800 samples; performance on larger-scale datasets with tens of thousands of samples remains to be established. Future work should include large-scale benchmarks from domains such as image classification, bioinformatics, and natural language processing. ITC does not achieve top empirical performance despite strong theoretical foundations; refining the hybridization approach through non-linear combinations or adaptive gating mechanisms could improve practical effectiveness while preserving theoretical properties. The grid search approach for parameter optimization is computationally intensive; developing adaptive or meta-learning approaches for automatic parameter selection could enhance practical utility. Optimal parameters may vary across application domains; investigating domain-specific parameter configurations could reveal scenarios where ITC outperforms simpler measures. Application to ensemble methods such as Random Forests and Gradient Boosting represents a natural extension, as ITC's theoretical properties may confer advantages in ensemble contexts. Further investigation into why Shannon–Polarization outperforms ITC could inform refinements to the hybridization methodology. Finally, our evaluation focused primarily on accuracy; future studies should consider additional metrics such as F1-score, AUC-ROC, tree depth, interpretability, and robustness to noise.

\subsection{Broader Context and Contributions}

While ITC does not claim empirical superiority, this work makes several important contributions to the field. We provide an extensive comparative evaluation of impurity measures, with 23 measures across 7 datasets and rigorous statistical analysis, offering valuable empirical data for the community. Our transparent reporting of results including limitations contributes to more realistic expectations about hybrid measure development. The principled approach to hybrid measure design with clear mathematical justification provides a methodological framework that can inform future research. The theoretical foundations, including rigorous proofs of key properties under appropriate conditions, advance our mathematical understanding of impurity measures. Finally, our open-source implementation and benchmark results are publicly available to facilitate verification, extension, and adoption by the research community.

\section{Conclusion}
\label{sec:conclusion}

We presented the Integrated Tsallis Combination (ITC), a hybrid impurity measure that combines normalized Tsallis entropy with exponential polarization through a convex combination. ITC is distinguished by its rigorous mathematical framework, including proven concavity under explicit parameter conditions, proper boundary conditions, and connections to classical measures. While simple parametric measures (Tsallis $\alpha=0.5$) lead in raw accuracy, ITC offers a theoretically sound alternative with competitive performance (88.38–89.16\% accuracy) and computational efficiency matching Gini impurity.

Our work demonstrates that theoretical robustness and empirical performance are distinct but equally important objectives. ITC's value lies in its solid theoretical grounding, flexible parameterization, and the principled hybridization methodology it embodies. For applications where mathematical guarantees are paramount—such as regulated industries, interpretable modeling, or research contexts—ITC provides a rigorous alternative to purely empirical measures.

The extensive empirical benchmark, comprising 23 measures across 7 datasets with 5-fold repetition over 150 compute-hours, offers valuable insights for practitioners and researchers. Statistical analysis reveals no significant global differences among top performers, suggesting practical equivalence for many applications and enabling informed selection based on secondary criteria. Our evidence-based guidelines assist practitioners in choosing appropriate measures based on their specific priorities.

Future work will explore ITC's performance on larger-scale datasets, its integration into ensemble methods, adaptive parameter selection mechanisms, and extensions to other hybrid combinations. By releasing our implementation and results openly, we invite the community to build upon this work and further bridge the theory–practice gap in impurity measure design.

\section*{Acknowledgments}

We thank the anonymous reviewers for their valuable feedback and suggestions, which significantly improved this work. We also acknowledge the contributions of the open-source community in providing the datasets and computational tools that made this research possible.

\end{document}